\documentclass[10pt,twocolumn,letterpaper]{article}

\usepackage[pagenumbers]{main}

\usepackage[table]{xcolor}
\usepackage{microtype}
\usepackage{graphicx}
\usepackage{booktabs}
\usepackage{amsmath}
\usepackage{amssymb}
\usepackage{mathtools}
\usepackage{amsthm}

\usepackage{color}
\usepackage{enumitem}
\usepackage{fontawesome5}
\usepackage{multirow}
\usepackage{makecell}

\usepackage{algorithm}
\usepackage{algpseudocode}   
\usepackage{etoolbox,siunitx}

\usepackage{tcolorbox}
\usepackage{caption}
\usepackage{hyperref}
\PassOptionsToPackage{nameinlink,capitalize}{cleveref}

\DeclareCaptionType{prompt}[Prompt]

\usepackage{kotex}

\definecolor{lblue}{RGB}{231, 66, 52}

\def\paperID{1}
\def\confName{RoboSense}
\def\confYear{IROS 2025}

\title{Robust Driving QA through Metadata-Grounded Context and Task-Specific Prompts \thanks{This manuscript is the technical report for The RoboSense Challenge 2025 (Track \#1: Driving with Language), describing our 4th-place solution.}}

\author{Seungjun Yu\thanks{Equally contributed}, \space Junsung Park\footnotemark[1], \space Youngsun Lim, \space Hyunjung Shim\thanks{Corresponding author}\\
\\
Korea Advanced Institute of Science and Technology\\
108, Taebong-ro, Seocho-gu, Seoul, Republic of Korea\\
{\small \texttt{\{seungjunyu, jshackist, youngsun\_ai, kateshim\}@kaist.ac.kr}}
}

\begin{document}

\maketitle

\begin{abstract}
We present a two-phase vision-language QA system for autonomous driving that answers high-level perception, prediction, and planning questions. In Phase-1, a large multimodal LLM (Qwen2.5-VL-32B) is conditioned on six-camera inputs, a short temporal window of history, and a chain-of-thought prompt with few-shot exemplars. A self-consistency ensemble (multiple sampled reasoning chains) further improves answer reliability. In Phase-2, we augment the prompt with nuScenes scene metadata (object annotations, ego-vehicle state, etc.) and category-specific question instructions (separate prompts for perception, prediction, planning tasks). In experiments on a driving QA benchmark, our approach significantly outperforms the baseline Qwen2.5 models. For example, using 5 history frames and 10-shot prompting in Phase-1 yields 65.1\% overall accuracy (vs. 62.61\% with zero-shot); applying self-consistency raises this to 66.85\%. Phase-2 achieves 67.37\% overall. Notably, the system maintains ≈96\% accuracy under severe visual corruption. These results demonstrate that carefully engineered prompts and contextual grounding can greatly enhance high-level driving QA with pretrained vision-language models.

\end{abstract}

\section{Introduction}
\label{sec:intro}
High-level question answering in autonomous driving requires interpreting complex, dynamic scenes and reasoning about future outcomes. For example, a system may be asked “Why is the pedestrian about to stop?” or “What maneuver will the red car perform next?” This demands connecting perception (object detection, state estimation), prediction (trajectory inference), and planning in a human-interpretable way. Vision-language models (VLMs) offer a promising framework for such queries, but existing work has limitations. Prior driving systems integrating language models (e.g., DriveGPT4~\cite{xu2024drivegpt4}, DriveVLM~\cite{tian2024drivevlm}, LMDrive~\cite{shao2024lmdrive}) have focused on end-to-end control or scene description, without explicitly answering arbitrary high-level questions. Conversely, driving QA benchmarks (e.g., DriveBench~\cite{xie2025vlms}, nuScenes-QA~\cite{qian2024nuscenes}) often target narrow subproblems (object querying, importance ranking) or rely on simple metrics (BLEU/ROUGE) that fail to capture reasoning quality. Moreover, VLMs are prone to hallucination and bias when faced with open-ended driving questions without structured guidance.

Visual question answering (VQA) in autonomous driving typically spans three core tasks: perception (e.g. identifying traffic lights, road signs), prediction (e.g. anticipating other agents’ movements), and planning (e.g. determining safe routes). Recent datasets address these components: for example, LingoQA~\cite{marcu2024lingoqa} uses video frames along with sensor metadata (speed, steering angle, weather) to generate QA pairs, and Box-QAymo~\cite{etchegaray2025box} constructs box-referenced QA tasks by extracting 3D object metadata from driving scenes. While these efforts enrich task diversity, most existing approaches use metadata only offline (e.g. for dataset creation) or assume idealized scene knowledge. In LingoQA, metadata helps formulate questions but is not directly used at inference time to ground answers. Box-QAymo provides a benchmark built on 3D annotations, but it does not explicitly feed this structured information into the model’s prompts. As a result, current VLM-based QA systems may still struggle with ambiguous scenes that lack explicit grounding. 
\begin{figure*}
  \centering
  \includegraphics[width=0.7\linewidth]{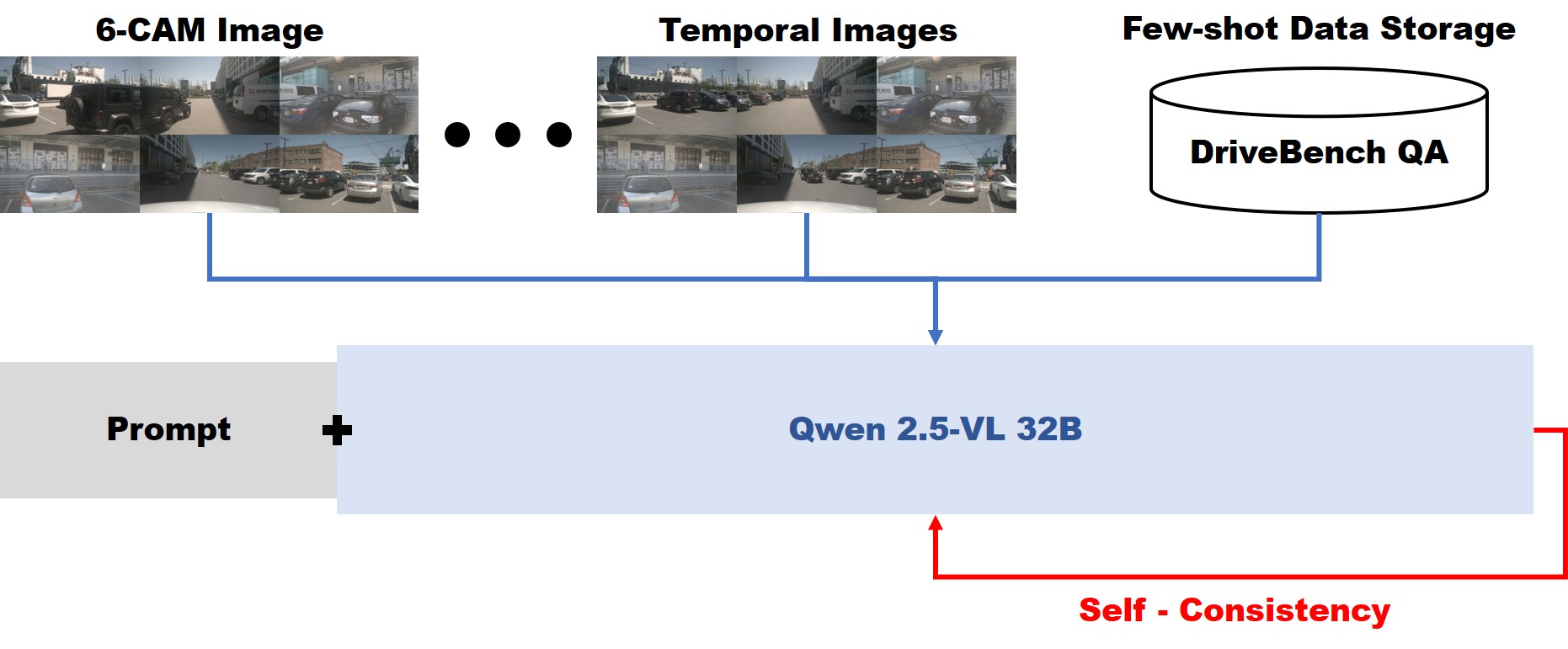}
  \caption{\textbf{Phase-1 overview.} Multi-view inputs (6-CAM) and a short temporal window of history frames are fused with a few-shot QA (DriveBench QA) into a domain-aware \emph{Prompt}, which conditions \textbf{Qwen~2.5-VL~32B}.}
  \label{fig:phase1}
\end{figure*}

In this paper, we address these gaps by directly incorporating structured driving metadata into the QA process. In Phase-1, we feed multi-view camera images and a brief history of past frames into a pretrained VLM, and use a carefully crafted prompt that follows a structured Perception→Prediction→Planning chain-of-thought. We incorporate a small set of in-context examples (5–10 shot prompting) to demonstrate the reasoning style. In addition, we use a self-consistency decoding strategy: the model generates multiple reasoning chains per query and we aggregate them by voting to improve accuracy. In Phase-2, we further enrich the input prompt with scene metadata (nuScenes annotations) and vehicle state information. For example, we serialize ego-vehicle speed and heading computed from past frames into text. We also design task-specific prompts: for “Perception” questions we highlight visual cues, while for “Planning” questions we emphasize road rules and traffic agents. These combined enhancements ground the VLM in domain knowledge and question context.

The main contributions of this paper are summarized as follows:
\begin{itemize}
\item \textbf{Domain-Grounded Context.} We leverage structured nuScenes metadata – including object annotations and ego-vehicle motion – to provide concrete scene context and anchor model responses.
\item \textbf{Multi-Modal Context Injection.} We embed both textual descriptions and visual cues (e.g. annotated 3D bounding boxes) of objects and ego-motion into prompts, enhancing model grounding and reducing spurious outputs.
\item \textbf{Task-Specific Prompting.} We design separate prompting strategies for perception, prediction, and planning queries, enabling the model to focus on relevant information for each task.
\end{itemize}

\begin{figure*}
  \centering
  \begin{subfigure}{0.64\linewidth}
    \centering
    \includegraphics[width=0.99\linewidth]{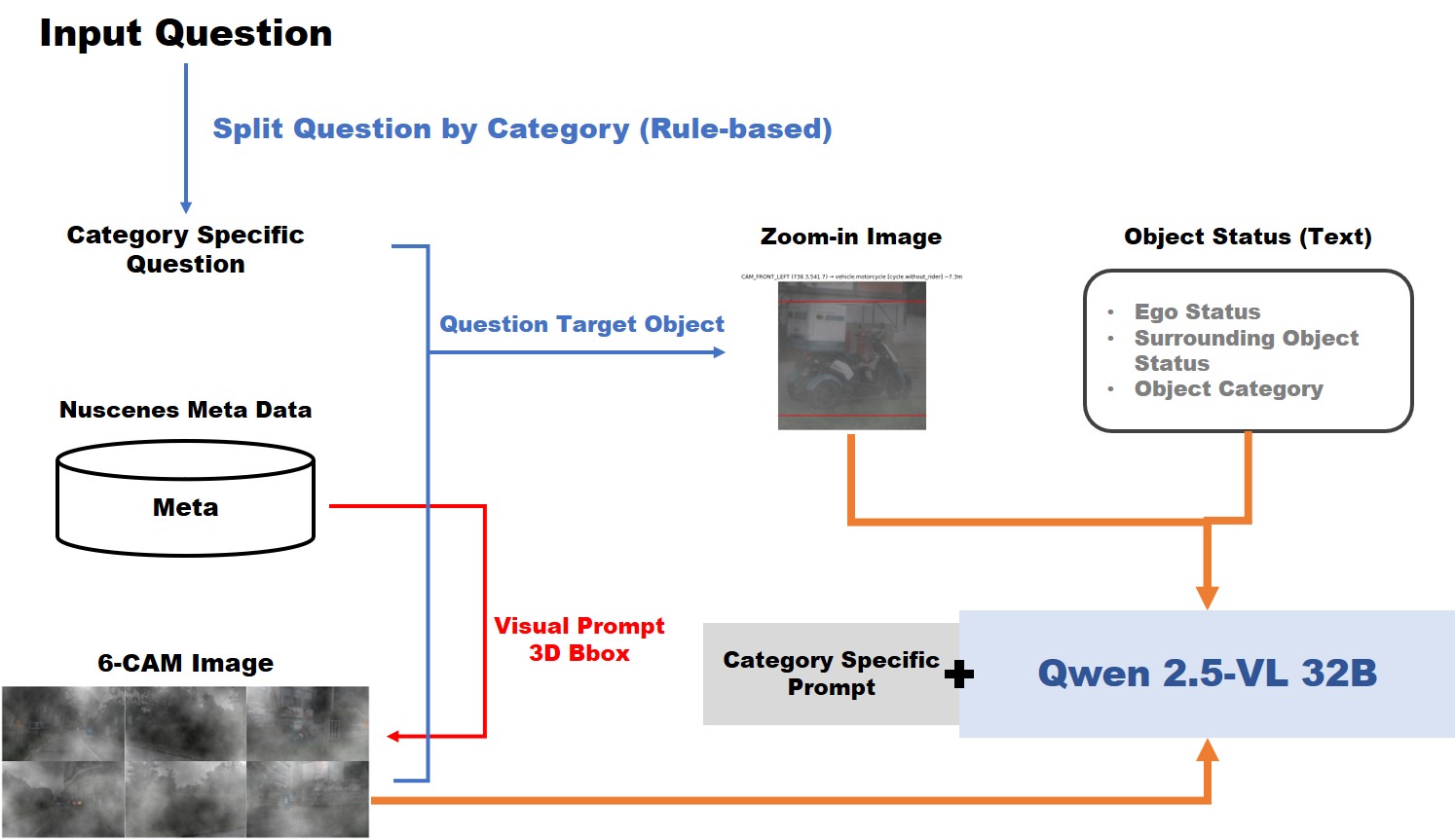}
    \caption{Phase-2 overview.}
    \label{fig:short-a}
  \end{subfigure}
  \hfill
  \begin{subfigure}{0.35\linewidth}
    \centering
    \includegraphics[width=0.9\linewidth]{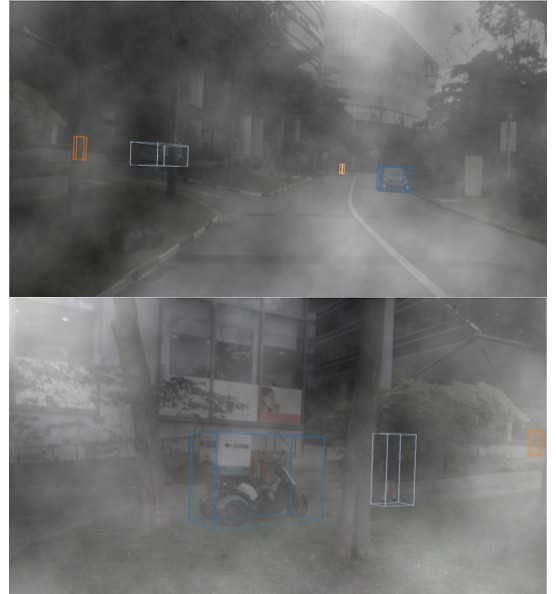}
    \caption{Example of Nuscenes Images with 3D bounding box visual prompt.}
    \label{fig:short-b}
  \end{subfigure}
  \caption{\textbf{Phase-2 overview.} A rule-based router splits the input question by category.
    NuScenes metadata (pose/map), zoom-in crops (question target), and object/ego status (text) are injected to form a task-specific prompt, then fed to \textbf{Qwen~2.5-VL~32B}.}
  \label{fig:phase2_design}
  \label{fig:phase2}
\end{figure*}
\section{Related Works}

The related works in this technical report can be categorized into three main areas:  
(A) autonomous driving using VLMs or LLMs,  
(B) language–driving combined datasets and benchmarks, and  
(C) reliability, robustness, hallucination, and evaluation metrics of VLMs.  

\subsection{Vision-Language-based autonomous driving}  
Attempts to integrate VLMs or LLMs into autonomous driving pipelines have evolved along two directions:  
(i) connecting perception–prediction–planning through interpretable high-level natural language interaction, and (ii) extending toward language–action (end-to-end) integration.  
DriveGPT4 presented an interpretable end-to-end framework that jointly produces natural language reasoning and low-level control quantities in response to video–text queries~\citep{xu2024drivegpt4}.  
DriveVLM structured a VLM-centric driving pipeline consisting of scene understanding, analysis, and hierarchical planning~\citep{tian2024drivevlm}. 
LMDrive applied LLMs to closed-loop driving control to explore stability in real driving scenarios~\citep{shao2024lmdrive}.  
In addition, studies such as OmniDrive~\citep{wang2025omnidrive} utilized counterfactual reasoning on the nuScenes dataset to tightly connect trajectory planning and language reasoning.  
These studies directly form the preceding context for the question types (Perception, Prediction, Planning) and the evaluation focus (reliable responses to high-level queries) in this challenge.  
However, these studies have not yet been validated on a QA set like the one used in this challenge.  

\subsection{Language-Driving Datasets and Benchmarks}  
Language–driving combined datasets originated from early tasks focused on “explanation” and “instruction/pointing.” More recently, they have expanded to include graph-structured, multi-view, and compositional queries.  
BDD-X~\citep{kim2018textual} annotated driving videos with explanations of planning actions to enhance interpretability regarding “why the vehicle drives that way”.  
Talk2Car~\citep{deruyttere2019talk2car} grounded natural language commands to object bounding boxes in the nuScenes dataset~\citep{caesar2020nuscenes}, aiming to improve driving scene understanding.  
nuScenes-QA~\citep{qian2024nuscenes} introduced a large-scale driving VQA benchmark based on the nuScenes dataset, enabling multi-view and multimodal questioning.  
Rank2Tell~\citep{sachdeva2024rank2tell} simultaneously requires ranking of important objects and providing reasoning, thereby measuring “importance inference” for accident-prevention contexts in driving scenes.  
LingoQA~\cite{marcu2024lingoqa} generates QA pairs using video frames combined with sensor metadata such as speed, steering angle, and weather conditions.  
Box-QAymo~\cite{etchegaray2025box} formulates box-referenced QA tasks by leveraging 3D object metadata extracted from driving scenes. 
In particular, DriveLM~\citep{sima2024drivelm} connects the full Perception–Prediction–Planning process through graph-based VQA and adopts GPT-based scoring for evaluation.
This line of work enables high-level question answering on multi-view camera scenes, as in this track, and serves as the starting point for designing question formats (multiple-choice, open-ended, visual-grounded) and evaluation metrics. 

\subsection{Reliability and Robustness of VLMs}
In safety-critical environments, the reliability of VLMs hinges on mitigating data distribution bias, visual degradation (corruption) sensitivity, and hallucination generation.  
\textbf{DriveBench}~\citep{xie2025vlms} serves as a benchmark covering 17 settings—including clean, corrupted, and text-only scenes—across four driving tasks (Perception, Prediction, Planning, and Behavior), highlighting that VLMs often produce “plausible” answers even without proper visual grounding.
For instance, in the behavior MCQ of DriveLM-nuScenes, the answer distribution is heavily biased toward \emph{Going Straight} (about 78.6\%), leading to high accuracy even without visual input.  
It also points out that GPT-based evaluations can be distorted without clear context or rubrics.  
Beyond driving scenes, \citet{li2025r} proposed the multimodal robustness benchmark R-Bench to quantitatively assess real-world corruption resilience of MLLMs.  
Furthermore, \citet{guan2024hallusionbench} introduced HALLUSIONBENCH, a fine-grained benchmark for diagnosing errors in LVLMs along two dimensions: language hallucination and visual illusion.

\begin{figure*}[ht!]
\centering
\begin{tcolorbox}[title=\textbf{Anchor and Context Information}]
\textbf{=== Anchor Info for Question Objects ===}\\
CAM\_FRONT\_LEFT (738.3,541.7) $\rightarrow$ vehicle.motorcycle [cycle.without\_rider] $\sim$7.3m

\medskip 

\textbf{=== Full Anchor Context ===}\\
SCENE CONTEXT:\\
OBJECT CANDIDATES:\\
\textless CAM\_FRONT,1139.0,529.9\textgreater\ vehicle.car [vehicle.stopped] ($\sim$28.2 m)\\
\textless CAM\_FRONT,988.3,501.5\textgreater\ human.pedestrian.adult [pedestrian.moving] ($\sim$81.1 m)\\
\textless CAM\_BACK\_LEFT,521.3,488.9\textgreater\ human.pedestrian.adult [pedestrian.moving] ($\sim$12.5 m)\\
\textless CAM\_BACK\_LEFT,595.4,442.5\textgreater\ vehicle.car [vehicle.moving] ($\sim$55.2 m)\\
\textless CAM\_BACK,783.2,512.7\textgreater\ vehicle.car [vehicle.moving] ($\sim$27.9 m)\\
\textless CAM\_BACK,1274.0,477.1\textgreater\ human.pedestrian.adult [pedestrian.sitting\_lying\_down] ($\sim$36.8 m)\\
\textless CAM\_BACK,1012.4,493.3\textgreater\ human.pedestrian.adult [pedestrian.standing] ($\sim$36.8 m)\\
\textless CAM\_BACK,1020.2,496.6\textgreater\ human.pedestrian.adult [pedestrian.sitting\_lying\_down] ($\sim$38.6 m)\\
\textless CAM\_BACK,734.1,499.2\textgreater\ vehicle.car [vehicle.moving] ($\sim$99.1 m)\\
\textless CAM\_FRONT\_LEFT,744.6,537.8\textgreater\ vehicle.motorcycle [cycle.without\_rider] ($\sim$7.3 m)\\
\textless CAM\_FRONT\_LEFT,1144.6,512.3\textgreater\ human.pedestrian.adult [pedestrian.moving] ($\sim$10.0 m)\\
\textless CAM\_FRONT\_LEFT,1565.9,436.2\textgreater\ human.pedestrian.adult [pedestrian.standing] ($\sim$30.7 m)
\end{tcolorbox}
\captionof{prompt}{Example of anchor information and full scene context from meta data.}
\label{code:anchor_context}
\end{figure*}

\section{Method}
~\label{method}
Our solution consists of two stages: an initial Phase-1 deployment with few-shot prompting, and an enhanced Phase-2 system with rich context input and zero-shot prompting.
Both phases use the Qwen2.5-VL model as the core.
Figure~\ref{fig:phase1} illustrates the architecture details of Phase-1, and Figure~\ref{fig:phase2} presents those of Phase-2.

\subsection{Phase1}
\noindent\textbf{History Frame.}
In autonomous driving scenarios, providing the model with temporal context (multiple history frames) significantly improves understanding of dynamic elements. 
By incorporating a short sequence of past frames, the model can infer object motions and trends (e.g. a pedestrian’s trajectory or a vehicle’s turning movement) instead of relying on a single snapshot.
This temporal information is crucial because most objects in AD scenes are moving, and their future behavior depends on their motion history. 
However, there are diminishing returns beyond a certain number of frames – too many history frames inflate inference time without proportional gains in accuracy. 
Thus, we use an optimal short history window that captures key motion cues while keeping the model efficient. 
This strategy ensures the model sees enough to judge movements, but avoids needless overhead from excessive frames.

\medskip
\noindent\textbf{Chain of Thought.}
We employ chain-of-thought (CoT) prompting to encourage the model to reason through driving decisions in a step-by-step manner.
Instead of directly outputting an answer, the model is prompted to first perceive the object (perception), then predict its action (prediction), and finally plan the ego action (planning).
This process has been shown to significantly enhance decision-making performance and transparency in complex tasks.
In fact, Waymo’s recent end-to-end driving model EMMA\cite{hwang2024emma} reported a improvement in planning accuracy by using chain-of-thought reasoning, while also providing human-interpretable rationales for its decisions. 
By mimicking human-like reasoning, CoT helps break down the driving problem (which often involves multiple factors like other agents, traffic rules, and future predictions) into intermediate conclusions. We found that instructing the model to articulate its reasoning not only improved correctness but also made its behavior more explainable. 
In summary, chain-of-thought prompting elevates the model from a black-box to a transparent problem-solver, yielding more robust and verifiable outputs in our driving tasks.

\medskip
\noindent\textbf{Few-shot Learning.}
To further boost performance, we leverage few-shot in-context learning by providing the model with a handful of example scenarios and solutions in the prompt.
These examples act as demonstrations, showing the model how to reason about driving questions.
We observed on the DriveBench toolkit that giving a couple of driving QA examples significantly improved the model’s responses on new, unseen scenarios.
This aligns with general LLM findings that a few well-chosen exemplars can prime the model effectively. 
In our case, we include 5–10 formatted examples (e.g. a question about a traffic situation and a step-by-step answer) before asking the model to solve the actual query.
Instead of being chosen for general aspects of driving reasoning, the examples are selected according to categories, ensuring that similar questions are grouped consistently.
The model then continues in the established pattern for the new input.
This few-shot prompt tuning provided a notable performance gain – it essentially teaches the model the style of reasoning and desired output format by example, which helped especially in complex or edge-case scenarios where zero-shot answers were initially flawed.

\medskip
\noindent\textbf{Self-consistency.}
We also apply the self-consistency decoding strategy to enhance reliability.
In this approach, the model generates multiple independent reasoning paths (by sampling with different randomness) for the same prompt, and we aggregate the results to find the most consistent answer.
In practice, we prompt the model several times for each query (each time it may produce a slightly different explanation and answer), then use a majority vote or confidence heuristic to decide on the final output.
This simple ensembling of reasoning chains greatly improves accuracy, as also noted in prior research.
The intuition is that while any single run might make an error in reasoning, it’s unlikely that multiple runs will converge on the same wrong answer if we inject diversity.
Indeed, we found that the agreement of multiple reasoning samples often pointed to the correct solution, filtering out occasional logic mistakes.
The trade-off is extra computation – generating, say, five answers instead of one – but we mitigate this by keeping the number of runs modest (e.g. 5) such that we get a boost in performance without unacceptable latency.
Overall, self-consistency made our system’s outputs more robust and mission-critical reliable, since the final answer has effectively been cross-validated by the model’s own different “thoughts”.

\subsection{Phase2}

\noindent\textbf{nuScenes Meta Data \& Visual Evidence.}
We leverage nuScenes meta data to expose key scene information to the LLM via both text and visual prompts. All meta data are retrievable through the question’s \texttt{\detokenize{scene_token}}, enabling consistent, token-addressable access to the underlying logs and annotations.

\medskip
\noindent\textbf{Ego Status via History Frames.}
As essential meta data, we estimate the ego-vehicle state from short history frames and provide it to the LLM as concise text. By comparing consecutive ego poses, we compute speed, heading, and acceleration/deceleration, then serialize them into human-readable sentences with rounding and consistent units, e.g., \texttt{“Ego-vehicle speed: 8 m/s, accelerating; Ego heading: north-east (turning right).”} Rather than passing full temporal image stacks to the LLM, we precompute these signals offline and pass only the textual summary. This reduces computational load while giving the model explicit motion context it would otherwise have to infer from raw visuals.

\medskip
\noindent\textbf{3D Bounding Boxes as Visual Prompt and Text Prompts for Surrounding Objects.}
In Phase-2, corrupted images make object perception challenging. We therefore exploit meta data to render 3D bounding boxes on the original images, supplying the LLM with geometry-aware visual evidence and improving downstream reasoning (see figure~\ref{fig:short-b}). To further reduce ambiguity, we accompany the visuals with a structured textual description of surrounding objects (categories, relative positions, and salient attributes), as illustrated in Prompt ~\ref{code:anchor_context}.

\medskip
\noindent\textbf{Anchor-Centered Zoom-In for the Question Object.}
Recent findings indicate that presenting a zoomed-in view of the target object helps LLMs focus and improves recognition and reasoning\cite{hu2024visual}. Using meta data, we localize the question’s target object. Because the meta data and the question reference may not match perfectly, we apply a tolerant matching strategy to select the best candidate. We then draw a 2D bounding box on the image, crop a zoomed-in patch centered on the anchor object, and provide both the global context and the magnified crop to the LLM. This anchor-centered presentation empirically improves object grounding and answer consistency.

\medskip
\noindent\textbf{3-Step Reasoning.}
We implement a structured three-step reasoning process in the model’s chain of thought, reflecting the classic perception–prediction–planning trifecta in autonomous driving.
In the first step, the model performs a Scene Description: it lists the relevant objects, actors, and environmental factors in the current scene (e.g. vehicles, pedestrians, traffic signals, road layout) and notes their key attributes or states.
Next, the model conducts Scene Analysis, where it examines the relationships and potential interactions – for instance, assessing the intentions or future motions of other agents (is that car ahead about to change lanes? is the pedestrian intending to cross?) and how they might affect the ego vehicle.
It also considers any risks or rule compliance issues at this stage (such as distances closing in too fast, or who has right of way). 
Finally, the model moves to Planning/Decision: here it synthesizes the insights from prior steps to formulate the appropriate driving decision or answer to the query.
This could mean deciding to slow down, identifying that “yes, the ego vehicle should yield,” or whatever action/answer is warranted, and explaining the rationale behind it.
We prompt the model to explicitly follow these three steps in its answer – effectively making the chain-of-thought modular. This approach was inspired by DriveVLM’s reasoning modules and we found it very effective.
By breaking the reasoning into scene description → analysis → plan, we ensure the model doesn’t skip over any layer of logic.
Each step feeds into the next, and errors can be caught in intermediate steps.
The overall effect is a coherent, well-structured answer that first paints the scene, then evaluates it, then concludes with a decision.
This three-step CoT not only improved correctness (since the model methodically considers all factors) but also made the outputs inherently explainable – mirroring how a human driver might justify their actions by saying, “I see X, it might do Y, so I will do Z.”

\medskip
\noindent\textbf{Category Specific Prompt.}
We argue that, in driving scenes, a single prompt or unified method cannot handle all question categories. 
In particular, perception, prediction, and planning are interrelated yet require substantially different reasoning skills, evidence, and output formats.
Therefore, we explicitly split questions by category and apply tailored prompts and procedures per task.
We implement a lightweight router that detects the category and dispatches to a dedicated template:
\begin{itemize}
  \item \textbf{Perception} \textrightarrow{} object/state grounding: enumerate salient actors, spatial relations, and visible signals; favor concise, evidence-first answers.
  \item \textbf{Prediction} \textrightarrow{} short-horizon intent/trajectory reasoning: condition on recent history and right-of-way cues; require probabilistic or most-likely outcomes.
  \item \textbf{Planning} \textrightarrow{} rule- and safety-aware decision making: instruct the model to weigh risks, traffic laws, and comfort; produce actionable, unambiguous commands.
\end{itemize}

These category-specific prompts act as conditional knowledge injection, steering the model toward the most relevant priors and checks for each question type.
In practice, this removes the one-tone-fits-all behavior, improves answer accuracy and context awareness, and yields more reliable driving decisions across diverse scenarios.

\begin{figure*}[ht!]
\centering
\begin{tcolorbox}[title=\textbf{System Prompt}]
\scriptsize
You are an autonomous driving assistant.\\

\textbf{TASK DESCRIPTION:}\\
In this phase, participants must answer high-level driving questions on possibly corrupted images.\\
You must rely on anchors and provided scene context.\\
Six camera views: [CAM\_FRONT, CAM\_FRONT\_RIGHT, CAM\_FRONT\_LEFT, CAM\_BACK, CAM\_BACK\_RIGHT, CAM\_BACK\_LEFT].\\
Coordinates: $<$id, camera\_view, x, y$>$ with resolution 1600$\times$900.\\

\textbf{ANCHORS:}\\
- The provided ANCHORS are the *only valid set of objects* in the current scene.\\
- No additional objects exist beyond this anchor list; do not hallucinate.\\
- Always ground your reasoning and final answers explicitly in the anchor information (e.g., category, distance, position, status, occlusion).\\
- When $<$cID,CAM,x,y$>$ tokens are referenced in the question, match them with the closest or overlapping anchor and describe reasoning.\\

\textbf{RULE:}\\
For every question (MCQ, perception, planning, prediction), always use exactly three sections:\\
1. Observations: factual description using anchors.\\
2. Reasoning: step-by-step logic with anchors.\\
3. Answer: final result.\\
- For MCQs: The Answer MUST repeat both the letter and the option text.\\
- For others: short factual sentence or action.\\
- Each Observations and Reasoning must have $\geq$2 sentences.\\
- Always weave anchor info ($\sim$distance m, category) into Observations/Reasoning.\\
- Never skip sections.\\
- Use only controlled vocabulary (adult, car, truck, stopped, moving, etc.).\\

\textbf{OUTPUT RULES (CRITICAL):}\\
- Vocabulary only: adult, child, car, truck, bus, bicycle, motorcycle, cyclist, pedestrian, left, center, right, moving, stopped, turning-left, turning-right, merging, braking, occluded, clear.\\
- Distances: integer meters or $\sim<$int$>$ m. Units: ``m''.\\
- No hedging (avoid: maybe, likely, seems).\\
- Answers must be long, detailed, and descriptive, weaving anchor information into complete sentences.\\
- Both Step 1 and Step 2 should be multi-sentence outputs, not short phrases.\\
- Do not use quotes or markdown styling in the final answer.
\end{tcolorbox}
\captionof{prompt}{System prompt rules and structured answering format.}
\label{prompt:system_prompt}
\end{figure*}
\begin{figure*}[ht!]
\centering
\begin{tcolorbox}[title=\textbf{Driving Domain Knowledge}]
\scriptsize
You are assisting vision-language autonomous driving reasoning.\\
Always follow the structured stack: Perception $\rightarrow$ Prediction $\rightarrow$ Planning $\rightarrow$ Action.\\

\textbf{PERCEPTION:}\\
- Detect and classify lane geometry: lane count, curvature, merges, and splits.\\
- Identify road controls: signals, stop/yield signs, lane markings, crosswalks, speed bumps, traffic cones, barriers.\\
- Recognize and categorize road users: vehicles, pedestrians, cyclists, motorcycles, buses, trucks.\\
- Estimate distances in meters, relative positions (left, center, right), and states (moving, stopped, braking, turning, merging).\\
- Consider occlusions from large vehicles, curves, or infrastructure.\\

\textbf{PREDICTION:}\\
- Infer future motion from current kinematics (speed, heading, turn signals, brake lights).\\
- Pedestrians: anticipate intent to cross when near crosswalks or curbs, especially if facing the road.\\
- Vehicles: detect lane-change or merging cues from wheel angle, lateral drift, or signal lights.\\
- Heavy vehicles (trucks, buses): longer stopping distance, wider turning radius, blind spots.\\
- Cyclists/motorcyclists: higher lateral movement uncertainty, frequent lane filtering.\\
- Always evaluate risk of collision in the next 3–5 seconds.\\

\textbf{PLANNING:}\\
- Maintain safe time headway ($\geq$2s normal, $\geq$3s in poor weather or congestion).\\
- Prioritize vulnerable road users (pedestrians, cyclists) over maintaining speed.\\
- Obey traffic signals and right-of-way rules strictly.\\
- Avoid unnecessary lane changes; only change when safe and beneficial.\\
- Smooth maneuvers preferred: reduce abrupt braking, keep steering gradual.\\
- Keep clear of occluded zones (e.g., behind parked cars or large trucks).\\
- Prepare fallback strategies: if uncertainty is high, slow down and wait.\\

\textbf{ACTION RULES OF THUMB:}\\
- If a lead vehicle brakes, increase following gap and prepare to stop smoothly.\\
- If a pedestrian/cyclist is approaching or waiting near a crosswalk, slow down and yield.\\
- At intersections: check all directions, anticipate merging/turning vehicles, and avoid blocking.\\
- In multi-lane roads: watch for merging traffic from ramps and lane changes from adjacent lanes.\\
- In poor visibility (fog, night, rain): reduce speed, increase following distance, and focus on road markings.\\
- On highways: keep right unless overtaking, maintain steady speed, avoid unnecessary lane weaving.\\
- In narrow roads: anticipate parked cars opening doors, pedestrians stepping off curbs.\\
- Always prioritize safety and legality over efficiency or speed.
\end{tcolorbox}
\captionof{prompt}{Driving domain knowledge used for autonomous driving reasoning.}
\label{prompt:domain_knowledge}
\end{figure*}
\begin{table*}[ht]
\centering
\small
\setlength{\tabcolsep}{4pt}
\begin{tabular}{ccccccccc}
\hline
\textbf{Temporal} & \textbf{Few-shot} & \textbf{Percep.-MCQs} & \textbf{Percep.-Obj} & \textbf{Percep.-Scene} & \textbf{Prediction} & \textbf{Plan.-Scene} & \textbf{Plan.-Obj} & \textbf{Overall} \\
\hline
\rowcolor{gray!15}
\multicolumn{9}{c}{\textbf{Qwen2.5-VL-7B}}\\
\rowcolor{gray!15}
\multicolumn{2}{c}{\textbf{Baseline}}&75.5&29.2&22.2&59.2&29.6&31.2&42.5\\
\hline
0&5&54.72&28.63&28.67&59.2&38.68&36.95&44.32\\
0&10&64.15&30.64&30.55&59.2&43.38&37.24&45.74\\
5&10&71.7&26.09&36.25&59.2&48.53&40.22&46.85\\
\hline
\rowcolor{gray!15}
\multicolumn{9}{c}{\textbf{Qwen2.5-VL-32B}}\\
\hline
5  & 0  & 86.79 & 55.45 & 61.72 & 61.30 & 65.48 & 63.60 & 62.34 \\
5  & 10 & \underline{90.57} & \underline{60.53} & 60.70 & \underline{65.52} & \underline{69.70} & 62.46 & \underline{65.12} \\
5  & 15 & 84.91 & 59.80 & \underline{62.73} & 65.33 & 69.22 & 60.91 & 64.36 \\
7  & 10 & 84.91 & 61.33 & 56.56 & 63.60 & 67.43 & 60.12 & 63.28 \\
10 & 0  & 86.79 & \textbf{62.05} & 55.70 & 60.92 & 65.00 & \textbf{61.97} & 62.61 \\
10 & 5  & \underline{90.57} & 59.94 & 57.42 & 59.58 & 64.04 & 60.25 & 61.41 \\
10 & 10 & 77.36 & 60.10 & 54.61 & 63.79 & 63.53 & 60.87 & 62.49 \\
\midrule
\rowcolor{gray!15}
\multicolumn{2}{c}{\textbf{Self-Consistency (5,10)}} & \textbf{94.34} & 58.81 & \textbf{68.67} & \textbf{67.43} & \textbf{72.31} & \underline{64.45} & \textbf{66.85} \\
\hline
\end{tabular}
\caption{\textbf{Results of Phase 1}: Performance comparison with \underline{Qwen2.5-VL-7B} and \underline{Qwen2.5-VL-32B}. The best score in each column is shown in \textbf{bold}, and the second-best is \underline{underlined}.}
\label{tab:temporal_fewshot_results}
\end{table*}

\begin{table*}[ht]
\centering
\small
\setlength{\tabcolsep}{4pt}
\resizebox{\textwidth}{!}{%
\begin{tabular}{cccccccccccc}
\toprule
\textbf{Visual Prompt} & \textbf{Object Meta} & \textbf{Ego Status} & \textbf{Task-specific Prompt} &
\textbf{Percep.-MCQ} & \textbf{Percep.-Obj} & \textbf{Percep.-Scene} & \textbf{Prediction} &
\textbf{Plan.-Scene} & \textbf{Plan.-Obj} & \textbf{Corruption} & \textbf{Overall} \\
\midrule
\rowcolor{gray!15}
\multicolumn{4}{c}{\textbf{Baseline (Qwen2.5-VL-7B)}} 
 & 75.51 & 25.54 & 17.42 & 61.56 & 32.67 & 28.04 & 23.08 & 40.63 \\
\rowcolor{gray!15}
\multicolumn{4}{c}{\textbf{Baseline (Qwen2.5-VL-32B)}} 
 & 68.37 & 39.20 & 45.89 & 62.02 & 52.05 & 43.35 & 82.69 & 52.82 \\
\midrule
No  & No  & Yes & Normal                  & 70.41 & 30.98 & 55.64 & 64.34 & 60.50 & \textbf{65.10} & \textbf{100} & 56.94 \\
Yes & Yes & Yes  & Normal                  & 74.49 & 38.43 & 48.75 & 68.76 & 56.62 & 49.84          & \textbf{100} & 58.22 \\
Yes & Yes & Yes & Perception-low           & \textbf{96.94} & 37.97 & 53.91 & 64.81 & 57.44 & 52.40 & 97.12 & 58.48 \\
Yes & Yes & Yes & Perception-high    & \underline{91.84} & \underline{42.79} & 63.92 & 75.96 & 61.26 & \underline{60.73} & 97.12 & \underline{63.26} \\
Yes & Yes & Yes & Prediction               & 66.33 & 35.70 & \textbf{65.20} & \underline{80.02} & 60.95 & 48.00 & 62.50 & 60.44 \\
Yes & Yes & Yes & Planning\&Corruption     & \underline{91.84} & 41.19 & 62.42 & 79.21 & \underline{61.35} & 51.96 & \textbf{100} & \underline{64.64} \\
\midrule
\rowcolor{gray!15}
Yes & Yes & Yes & Task Specific QA\&Prompt & \textbf{96.94} & \textbf{42.86} & \underline{64.04} & \textbf{85.37} & \textbf{61.46} & 51.96 & \textbf{100} & \textbf{67.37} \\
\bottomrule
\end{tabular}}
\caption{\textbf{Results of Phase 2}: Performance comparison with \underline{Qwen2.5-VL-32B}. Baselines are highlighted in gray. The best score in each column is shown in \textbf{bold}, and the second-best is \underline{underlined}.}
\label{tab:phase2_results}
\end{table*}

\section{Experiments}

\subsection{Experimental Setups}
~\label{experiment setup}
Our system is implemented using the Qwen-2.5-VL model as the core engine, deployed in an inferenceonly manner on a server with 4 x NVIDIA A6000 GPUs. 
The large 32B-parameter model demands substantial memory, so we distribute the multi-modal computation across GPUs for efficiency. 
We also evaluated the smaller 7B variant on a subset of queries to gauge the effect of model scale , though all final results use the 32B model for its superior accuracy. 
Each query (one driving question with associated images) is processed in a few seconds, which is acceptable for offline evaluation (though not real-time).

\subsection{Implementation Details}
\noindent\textbf{System prompt.} The system prompt defines the agent’s role, scene assumptions, and answer format. It restricts the model to the provided anchor list~\ref{code:anchor_context} (preventing hallucinated objects) and enforces an Observations$\rightarrow$Reasoning$\rightarrow$Answer template for every response. It also narrows the vocabulary to a controlled driving lexicon (e.g., \textit{vehicle}, \textit{pedestrian}, \textit{stopped}, \textit{turning right}) and standardizes factual reporting (distances in meters, rounded integers). By compelling step-wise reasoning grounded in anchors, the prompt improves both consistency and factuality; in practice, we observe fewer hallucinations and higher accuracy as the model follows a constrained reasoning chain. See Prompt~\ref{prompt:system_prompt} for the full template.

\medskip
\noindent\textbf{Driving knowledge} This prompt summarizes domain rules and best practices, organized along the Perception $\rightarrow$ Prediction $\rightarrow$ Planning $\rightarrow$ Action stack and highlighting key heuristics (lane geometry, traffic signals, agent kinematics, safety priorities). Prepending these guidelines biases the model toward realistic driving logic: it classifies lanes and road users more reliably, forecasts motion consistent with signals and right-of-way, and proposes maneuvers with safe headways (e.g., \emph{yield to pedestrian on crosswalk}). Empirically, adding driving knowledge improves planning and prediction performance. In short, the system prompt structures answers and grounds them in anchors, while driving knowledge supplies domain priors that steer reasoning toward plausible, safety-aware decisions. Details are provided in Prompt~\ref{prompt:domain_knowledge}.

\subsection{Comparative and Ablation Study}
~\label{main results}
We evaluate the impact of prompting strategies and contextual information as summarized in Table ~\ref{tab:temporal_fewshot_results} and Table ~\ref{tab:phase2_results}. In Phase-1 (Table~\ref{tab:temporal_fewshot_results}), we vary the number of few-shot examples and the length of the history window. Using a small number of in-context examples consistently improved performance: for instance, with 5 history frames, providing 5-shot prompts raised Perception MCQ accuracy from 86.79\% to 90.57\% and boosted the overall score (Table~\ref{tab:temporal_fewshot_results}). However, adding too many examples (10 or 15) did not yield further gains and in fact often degraded accuracy. We also find that increasing the frame history from 5 to 10 provides little benefit; the best overall accuracy was achieved with only 5 frames. These results are consistent with reports that very long input contexts can saturate or even hurt a model’s accuracy due to irrelevant content~\cite{liu2023lost}. Importantly, the same trend holds for the smaller 7B model, indicating the \emph{scalability of our methods across model sizes.}. Starting from a 7B baseline overall of 42.5, 10-shot prompting without history improves the overall score to 45.74, and adding a short 5-frame history with 10-shot further raises it to 46.85 (Table~\ref{tab:temporal_fewshot_results}), \emph{demonstrating that our approach is model-scalable}: the same recipe yields consistent relative gains at both 7B and 32B.

Finally, we applied self-consistency decoding (sampling multiple outputs and voting) in Phase-1. With 5 frames and 10-shot prompting, self-consistency produced the highest overall score (66.85) in Table~\ref{tab:temporal_fewshot_results}. This substantial gain is in line with recent findings that aggregating multiple reasoning paths (“self-consistency”) markedly improves chain-of-thought accuracy~\cite{wang2022self}. In summary, Phase-1 ablations show that a few well-chosen in-context examples and multiple-sample decoding enhance performance, while overly long prompts or histories do not.

In Phase-2 (Table~\ref{tab:phase2_results}), we assess the effects of task-specific prompting, ego-vehicle state input, and visual metadata. Starting from the “Normal” prompt condition, we observe that tailoring the prompt to each subtask yields large improvements. For example, using a high-level perception prompt (“Perception-high”) significantly boosts perception and prediction scores and raises overall accuracy from 58.22 to 63.26. Likewise, a prediction-focused prompt (“Prediction”) dramatically increases the Prediction score (to 80.02) and overall performance (60.44). The combined “Task Specific QA\&Prompt” setting (which integrates all specialized cues) achieves the best overall result (67.37) in Table~\ref{tab:phase2_results}. These gains confirm that aligning the model’s instructions with the task domain greatly enhances its reasoning, consistent with the general principle that domain-specific prompts improve VLM/LLM accuracy.

All Phase-2 configurations include the ego-vehicle’s state (speed, heading, etc.) as concise text in the prompt. We found that providing this ego status is crucial for context: even the baseline “Normal” prompt assumed Ego Status=Yes and achieved 56.94 overall. Omitting ego-state information in preliminary tests led to lower accuracy, indicating that explicit knowledge of the vehicle’s own motion helps the model reason about future actions and perception (e.g., anticipating motion given current speed). In other words, ego status input grounds the model’s understanding of self-motion, which is important for tasks like prediction.

Finally, Table~\ref{tab:phase2_results} shows the contribution of visual metadata. Adding an explicit “Visual Prompt” (the vanishing point cue) and structured “Object Meta” descriptions produces modest gains over the text-only baseline. For example, comparing the “No Visual Prompt” vs. “Yes Visual Prompt” rows (both with ego status) shows an overall improvement from 56.94 to 58.22. Some perception and prediction scores improve with these cues (e.g. Perception MCQ increases), though planning scores saw mixed effects. Intuitively, giving the model explicit geometric context and a scene graph helps its visual reasoning. This is in line with findings that encoding scene-graph information (object and relation structure) improves VQA accuracy~\cite{teney2017graph}. Our results confirm that including structured visual information as part of the prompt provides additional context that the LLM can leverage, yielding a small but consistent performance gain in high-level driving tasks.


\begin{table*}[ht]
\centering
\small 
\setlength{\tabcolsep}{2pt} 
\begin{tabular}{lcccccccc}
\hline
\textbf{Method} & \textbf{Perception-MCQs} & \textbf{Percep.-Obj} & \textbf{Percep.-Scene} & \textbf{Prediction} & \textbf{Corruption-MCQs} & \textbf{Plan.-Scene} & \textbf{Plan.-Obj} & \textbf{Overall} \\
\hline
Baseline & 72.45 & 38.60 & 47.25 & 61.67 & 82.69 & 52.22 & 43.34 & 52.81 \\
VP (Text) & 69.39 & 41.92 & 41.75 & 62.49 & 95.19 & 38.62 & 37.64 & 51.08 \\
VP (Visual) & 76.53 & 40.67 & 45.04 & 66.78 & 96.15 & 29.69 & 34.18 & 51.10 \\
DG (Text) & 70.41 & 33.64 & 46.30 & 57.61 & 97.12 & 37.33 & 34.40 & 47.30 \\
DG (Visual) & 68.37 & 33.39 & 41.58 & 62.60 & 87.50 & 24.85 & 30.77 & 46.19 \\
VP+DG (Text) & 51.02 & 33.42 & 44.08 & 63.30 & 80.77 & 54.82 & 41.16 & 51.11 \\
VP+DG (Visual) & 54.08 & 34.11 & 46.88 & 68.87 & 83.65 & 54.34 & 40.64 & 53.37 \\
\hline
\end{tabular}
\caption{Comparison of different prompting strategies. VP = Vanishing Point, DG = Dominant Gradient Orientation. Metrics: Perception-MCQs, Prediction, and Corruption-MCQs are reported as Accuracy, while the others are reported as Weighted-VQA-Score. All experiments are done in Qwen-2.5VL-32B.}
\label{tab:results_failure_case}
\end{table*}

\section{Failure Cases}
\label{sec:failure_cases}

\subsection{Vanishing Point}
~\label{sec:vp}
If 3D understanding ability can be injected into VLMs, the performance on perception, prediction, and planning tasks in driving scenes can be further improved~\citep{lee20253d}.  
However, approaches such as those proposed by \citet{lee20253d}, which use correspondence methods like DUSt3R~\citep{wang2024dust3r}, require multi-view inputs and are computationally intensive.  
In driving scenes, it is common to have a situation where the vanishing point can be detected from a front-facing camera~\citep{guo2024vanishing}. 
Therefore, leveraging this property enables an efficient way to provide VLMs with 3D prior knowledge.

\subsubsection{Vanishing Point as Text Prompt}
~\label{sec:vp_text_prompt}
We guided the VLM to perform scene understanding while considering the vanishing point by providing a text-based instruction. 
The text prompt was given as shown in ~\cref{prompt:vp_text_prompt}.  

\begin{figure*}[ht!]
\centering 
\begin{tcolorbox}[title=\textbf{System Prompt}]
You are a helpful autonomous driving assistant that can answer questions about images and videos. 
You are providing images from multi-view sensors ordered as \texttt{[CAM\_FRONT, CAM\_FRONT\_RIGHT, CAM\_FRONT\_LEFT, CAM\_BACK, CAM\_BACK\_RIGHT, CAM\_BACK\_LEFT]}. 
The object coordinates are provided in the format of \texttt{<id, camera\_view, x, y>}. 
The coordinate is the center of the bounding box where the image resolution is 1600x900.\medskip 

\begin{itemize}
    \item \textbf{Orientation first} \\
    From the forward views (FRONT/FRONT\_LEFT/FRONT\_RIGHT), infer the road \textbf{vanishing point (VP)} and the \textbf{horizon line}. Some inputs may include a \textbf{visual prompt} on CAM\_FRONT: a yellow cross labeled “VP” (and a thin horizontal line). Treat it as a strong hint but verify with raw geometry.

    \item \textbf{Reasoning rule} \\
    Prefer geometric cues (VP/horizon/long edges/symmetry). If colors look corrupted, reason as if the image were grayscale. Use VP to infer lane directions, ego heading, and depth ordering.

    \item \textbf{Do not output} \\
    Do not output numeric VP/horizon values unless explicitly asked.

    \item First orient the scene: infer the horizon and the main VP from CAM\_FRONT.
    If a VP overlay (yellow cross) is present, treat it as a hint and verify with CAM\_FRONT\_LEFT/RIGHT. Use VP/horizon only for internal reasoning (do not output numeric values unless explicitly asked). If colors look unreliable, imagine the grayscale version
\end{itemize}
\end{tcolorbox}
\captionof{prompt}{System prompt for the Vanishing Point (VP) prompting.}
\label{prompt:vp_text_prompt}
\end{figure*}

For each task, we used task-specific prompts: for perception tasks, we used ``Answer with a single short sentence. If it is MCQ, output only the option letter.''; for prediction tasks, we used ``Focus on plausible near-term motion consistent with VP-aligned road geometry.''; and for planning tasks, we used ``Recommend the safest maneuver consistent with lane direction inferred from VP.''.  

As shown in~\cref{tab:results_failure_case}, all comparisons are reported relative to the baseline.  
When using only the vanishing point text prompt (row 2), Perception MCQ decreases to 69.39, which is -3.06 lower. 
Perception-object VQA increased +3.32 to 41.92.
Perception-scene VQA decreased -5.50 to 41.75.
Prediction increased +0.82 to 62.49.
Corruption robustness increased +12.50 to 95.19.
Planning VQA decreased. Scene level decreased -13.60 to 38.62, and object level decreased -5.70 to 37.64.
Overall score decreased -1.73 to 51.08.


\subsubsection{Vanishing Point as Visual Prompt}
~\label{sec:vp_visual_prompt}
\begin{figure}[ht]
    \centering
    \includegraphics[width=0.95\linewidth]{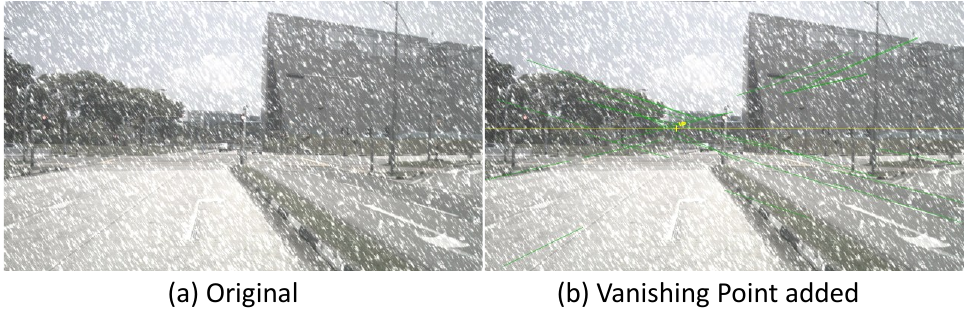}
    \caption{Visual prompts for the vanishing point on clean image. The yellow line represents the horizontal line, and the estimated vanishing point is marked with `VP`.}
    \label{fig:vp_clean}
\end{figure}

\begin{figure}[ht]
    \centering
    \includegraphics[width=0.95\linewidth]{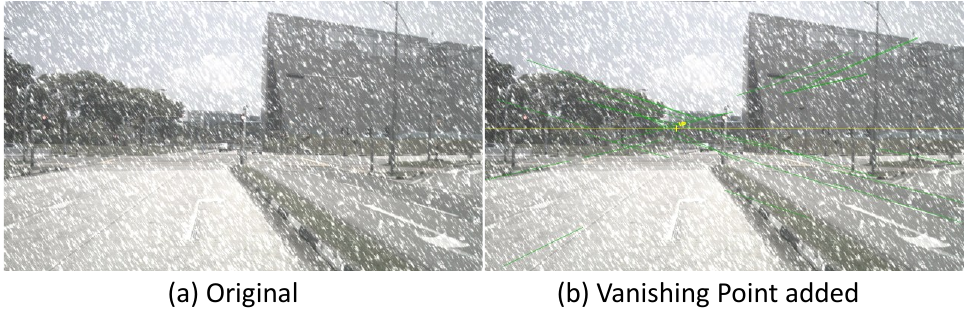}
    \caption{Visual prompts for the vanishing point on corrupted image. The yellow line represents the horizontal line, and the estimated vanishing point is marked with `VP`.}
    \label{fig:vp_cor}
\end{figure}

In addition to text prompting, we also experimented with injecting the vanishing point as a visual prompt.  
If the confidence score of the detected vanishing point was below 0.3, no vanishing point was provided.  
When the vanishing point was successfully detected, the corresponding visual prompts were generated as illustrated in~\cref{fig:vp_clean} and~\cref{fig:vp_cor}. 


When the vanishing point visual prompt was additionally used (row 3 of ~\cref{tab:results_failure_case}), Perception MCQ increased +4.08 to 76.53.
Perception-object VQA increased +2.07 to 40.67.
Perception-scene VQA decreased -2.21 to 45.04.
Prediction increased +5.11 to 66.78.
Corruption robustness increased +13.46 to 96.15.
Planning VQA decreased. Scene level decreased -22.53 to 29.69, and object level decreased -9.16 to 34.18.
Overall score decreased -1.71 to 51.10.

These results indicate that using vanishing points for inference improves perception and prediction performance but substantially degrades Planning VQA.  
Also, the effect of using the vanishing point as a visual prompt was not as strong as expected when compared to text prompting.  
This suggests that the language prior may dominate the model’s behavior, as also reported in previous work~\cite{chen2024mllm}.  

\subsection{Dominant Gradient Orientation}
~\label{sec:dgo}
Under image corruptions such as snow, rain, or foggy weather, much of the object edge information becomes obscured.  
If such edges are emphasized and their orientations are provided simultaneously, it becomes easier to understand the overall shape of objects.  
In particular, lanes and road boundaries are often occluded by corruption, despite being critical for determining drivable areas.    
Thus, emphasizing edge information while also providing orientation can help separate cars from roads and disentangle context, leading to improved scene understanding.  

\subsubsection{Dominant Gradient Orientation as Text Prompt}
~\label{sec:dgo_text_prompt}
To encourage the model to consider domain gradient orientation, we instructed it with the text prompt shown in ~\cref{prompt:dgo_text_prompt}.  

\begin{figure*}[ht!]
\centering 
\begin{tcolorbox}[title=\textbf{System Prompt}]
You are a helpful autonomous driving assistant that can answer questions about images and videos. 
You are providing images from multi-view sensors ordered as \texttt{[CAM\_FRONT, CAM\_FRONT\_RIGHT, CAM\_FRONT\_LEFT, CAM\_BACK, CAM\_BACK\_RIGHT, CAM\_BACK\_LEFT]}. 
The object coordinates are provided in the format of \texttt{<id, camera\_view, x, y>}. 
The coordinate is the center of the bounding box where the image resolution is 1600x900.\medskip

\begin{itemize}
    \item \textbf{Orientation-only rule} \\
    Estimate the scene’s \textbf{Dominant Gradient Orientation (DGO)} by computing Sobel gradients on a grayscale version of the forward views and building a weighted angle histogram ($\theta \in [0, \pi)$). Use the \textbf{mode of $\theta$} (the highest-mass bin) as the primary global direction cue. This cue is robust to color/brightness corruptions and helps with road heading/roll normalization and lane-alignment reasoning. \medskip

    \item \textbf{Visual Prompts} \\
    Some inputs may include an orientation visual prompt. Treat these as hints and still reason from the \textbf{global angle distribution}.
    \begin{itemize}
        \item ``[RGB | ORIENT]'' 2-panel (right = angle-colorized orientation map), or
        \item an RGB image with a semi-transparent orientation overlay, or
        \item the orientation map itself.
    \end{itemize} \medskip

    \item \textbf{How to read the orientation map (color legend)}
    \begin{enumerate}
        \item The map is HSV-coded from Sobel gradients: \textbf{H = $\theta$}, \textbf{S = 255}, \textbf{V $\propto \|\nabla I\|$} (gradient magnitude) normalized; thus
        \begin{itemize}
            \item \textbf{Hue (color)} encodes \textbf{gradient orientation $\theta$} modulo 180° (axial). \textbf{$\theta$ and $\theta+180$° look the same color.}
            \item \textbf{Value/brightness} encodes \textbf{edge strength}: brighter = stronger gradients; dark/near-black = weak or textureless regions.
        \end{itemize}
        \item Remember: hue is the \textbf{gradient direction}, which is \textbf{perpendicular to the visual edge direction}. Use this when reasoning about lines.
        \item Approximate hue→angle anchors (display-dependent):
        \begin{itemize}
            \item \textbf{Red $\approx$ 0°/180°} (left–right gradient), \textbf{Yellow/Green $\approx$ 30–60°}, \textbf{Cyan $\approx$ 90°} (up–down gradient), \textbf{Blue $\approx$ 120°}, \textbf{Magenta $\approx$ 150°}.
        \end{itemize}
        \item In \textbf{overlay} mode the hue tints the RGB image; focus on bright, edge-like regions rather than diffuse colors.
        \item The \textbf{yellow center line} is the global DGO estimate drawn through the image center; treat it as a high-level heading cue.
    \end{enumerate} \medskip

    \item When describing the map, use qualitative terms like ``mostly vertical/cyan gradients'' or ``mostly horizontal/red gradients'' unless a numeric angle is explicitly requested. \medskip

    \item Do \textbf{not} output numeric angles unless explicitly asked. \medskip
    
    \item First, estimate the Dominant Gradient Orientation (DGO) from CAM\_FRONT by computing Sobel gradients on a grayscale version and building a weighted $\theta$ histogram (0..$\pi$). Cross-check DGO consistency with CAM\_FRONT\_LEFT/RIGHT. Use DGO to stabilize heading/roll reasoning... and do not output numeric angles unless explicitly asked.
\end{itemize}

\end{tcolorbox}
\captionof{prompt}{System prompt for the Dominant Gradient Orientation (DGO) prompting.}
\label{prompt:dgo_text_prompt}
\end{figure*}

When only the Dominant Gradient text prompt was used (row 4 of ~\cref{tab:results_failure_case}), Perception MCQ decreased -2.04 to 70.41.
Perception-object VQA decreased -4.96 to 33.64.
Perception-scene VQA decreased -0.95 to 46.30.
Prediction decreased -4.06 to 57.61.
Corruption robustness increased +14.43 to 97.12.
Planning VQA decreased. Scene level decreased -14.89 to 37.33, and object level decreased -8.94 to 34.40.
Overall score decreased -5.51 to 47.30.


\subsubsection{Dominant Gradient Orientation as Visual Prompt}
~\label{sec:dgo_visual_prompt}
\begin{figure}[ht]
    \centering
    \includegraphics[width=0.95\linewidth]{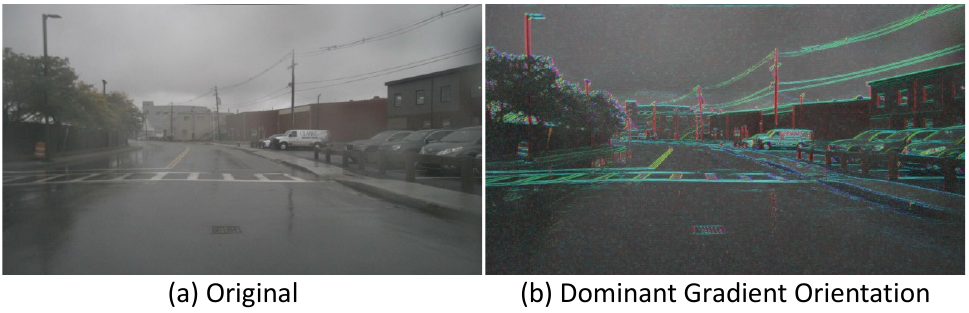}
    \caption{Visual prompts for the dominant gradient orientation on a clean image. Brighter regions indicate stronger gradients, darker regions indicate weaker gradients, and colors represent gradient directions.}
    \label{fig:dgo_clean}
\end{figure}

\begin{figure}[ht]
    \centering
    \includegraphics[width=0.95\linewidth]{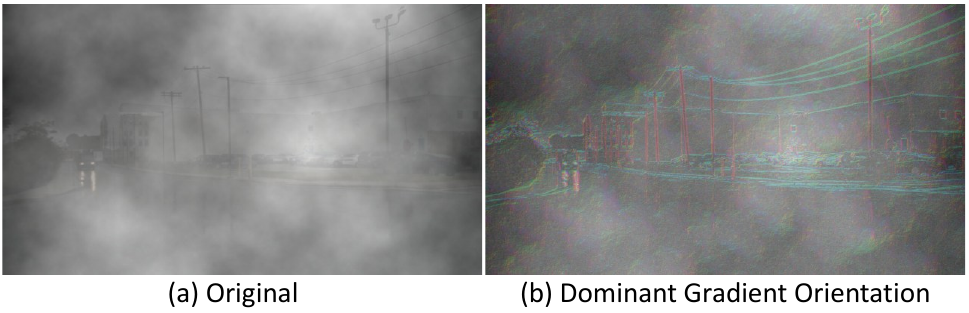}
    \caption{Visual prompts for the dominant gradient orientation on a corrupted image. Brighter regions indicate stronger gradients, darker regions indicate weaker gradients, and colors represent gradient directions.}
    \label{fig:dgo_cor}
\end{figure}

In addition to text prompting, we also experimented with injecting Dominant Gradient Orientation as a visual prompt.  
When the dominant gradient orientation was extracted and provided as a visual prompt, the resulting visualizations were obtained as illustrated in~\cref{fig:dgo_clean} and~\cref{fig:dgo_cor}.  


When only the Dominant Gradient visual prompt was used (row 5 of ~\cref{tab:results_failure_case}), Perception MCQ decreased -4.08 to 68.37.
Perception-object VQA decreased -5.21 to 33.39.
Perception-scene VQA decreased -5.67 to 41.58.
Prediction increased +0.93 to 62.60.
Corruption robustness increased +4.81 to 87.50.
Planning VQA decreased. Scene level decreased -27.37 to 24.85, and object level decreased -12.57 to 30.77.
Overall score decreased -6.62 to 46.19.
Thus, while leveraging Dominant Gradient Orientation yield substantial gains in corruption robustness, they reduce performance across other tasks.  
Similar to the vanishing point case, the effect of visual prompting was weaker than that of text prompting, suggesting that the strong language prior may dominate, consistent with prior findings~\citep{chen2024mllm}.  
Moreover, since prompts may not effectively capture color-related information~\citep{liang2025colorbench}, the resulting improvement might be only marginal.

\section{Difficulties during the Experiment}

\noindent\textbf{Methods are not Scalable}
Among the observations from the experiments in~\cref{sec:vp} and~\cref{sec:dgo}, we found that the two methods which showed improvements on 7B did not yield benefits on 32B.  
This may be due to an engineering mistake in parameter settings during model scaling, or it may result from the fact that when scaling from Qwen-7B to Qwen-32B, primarily the language encoder grows.  
In this case, the stronger language prior could overshadow the contribution of visual information, preventing the performance gains observed at the smaller scale.  
Detailed results are presented in ~\cref{tab:results_comparison_7b}.  

\begin{table*}[ht]
\centering
\small 
\setlength{\tabcolsep}{4pt} 
\resizebox{\textwidth}{!}{%
\begin{tabular}{lcccccccc}
\hline
\textbf{Method} & \textbf{Percep.-MCQs} & \textbf{Percep.-Obj} & \textbf{Percep.-Scene} & \textbf{Prediction} & \textbf{Corrupt.-MCQs} & \textbf{Plan.-Scene} & \textbf{Plan.-Obj} & \textbf{Overall} \\
\hline
7B - Baseline & 75.51 & 25.54 & 17.42 & 61.56 & 23.08 & 28.24 & 28.04 & 40.63 \\
7B - VP (Text) & 73.47 & 27.38 & 22.33 & 61.56 & 33.65 & 41.62 & 37.50 & 45.22 \\
7B - VP (Visual) & 74.49 & 27.50 & 19.15 & 61.56 & 33.65 & 47.16 & 38.06 & 45.82 \\
7B - DG (Text) & 72.45 & 32.20 & 21.67 & 61.56 & 31.73 & 44.53 & 45.96 & 48.33 \\
7B - DG (Visual) & 72.45 & 30.20 & 23.55 & 61.56 & 24.04 & 44.07 & 46.15 & 47.74 \\
\hline
\hline
32B - Baseline & 72.45 & 38.60 & 47.25 & 61.67 & 82.69 & 52.22 & 43.34 & 52.81 \\
32B - VP (Text) & 69.39 & 41.92 & 41.75 & 62.49 & 95.19 & 38.62 & 37.64 & 51.08 \\
32B - VP (Visual) & 76.53 & 40.67 & 45.04 & 66.78 & 96.15 & 29.69 & 34.18 & 51.10 \\
32B - DG (Text) & 70.41 & 33.64 & 46.30 & 57.61 & 97.12 & 37.33 & 34.40 & 47.30 \\
32B - DG (Visual) & 68.37 & 33.39 & 41.58 & 62.60 & 87.50 & 24.85 & 30.77 & 46.19 \\
\hline
\end{tabular}}
\caption{Comparison between 7B and 32B. VP = Vanishing Point, DG = Dominant Gradient Orientation. Metrics: Perception-MCQs, Prediction, and Corruption-MCQs are reported as Accuracy, while the others are reported as Weighted-VQA-Score. All experiments are done in Qwen-2.5VL-7B and Qwen-2.5VL-32B.}
\label{tab:results_comparison_7b}
\end{table*}

\noindent\textbf{Scoring Metric Manipulation with Text Length}
\label{sec:difficulty_scoring}
We observed that the scoring tends to assign higher final scores as the text length increases.  
As noted in~\citet{chen2024mllm}, when evaluation is performed with models such as GPT-4o, longer responses often receive better scores even under the same rubric.  
Consistently, we also found in our experiments that longer text responses were evaluated more favorably.  

\noindent\textbf{Problem on Evaluating MCQ}
A particular issue was observed in MCQ evaluation: when the model outputs simply ``A'' the answer is marked incorrect, while the output ``A. [Answer]'' is considered correct.  

\begin{figure*}[ht]
    \centering
    \includegraphics[width=0.95\linewidth]{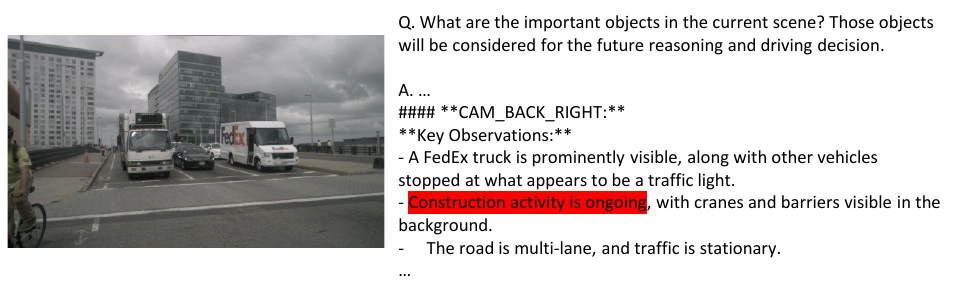}
    \caption{Hallucination case in driving scene VLMs. The model asserts a construction site in the rear left camera although no such scene is present. It likely reuses evidence detected in the front camera and projects it to the rear left view without verifying visual cues. This cross view leakage suggests strong language priors and weak visual grounding across views.}
    \label{fig:hallucination}
\end{figure*}
\begin{figure*}[ht]
    \centering
    \includegraphics[width=0.95\linewidth]{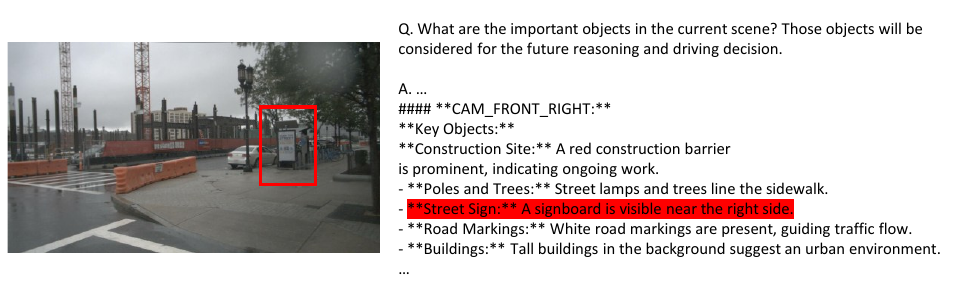}
    \caption{Wrong prediction case in driving scene VLMs. The model recognizes an advertising board commonly found at a bus stop as a street sign indicating a traffic signal.}
    \label{fig:wrong_output}
\end{figure*}

\section{Future Works}

\noindent\textbf{VLM Hallucination and Wrong Output due to Language Priors.}
As discussed by \citet{chen2024multi}, VQA in driving scenes often includes many objects, which makes the task more vulnerable to hallucination.  
In our analysis, we also observed cases where the model relied on language priors and produced answers without carefully inspecting the image.  
This behavior aligns with the observations reported by \citet{chen2024mllm}.  
As shown in ~\cref{fig:hallucination}, the model produces incorrect answers by asserting the presence of a non-existent construction environment. This illustrates a typical hallucination case where visual evidence does not support the claimed scene context. 
In addition, due to the strong language prior of the VLM, we observe in~\cref{fig:wrong_output} that an advertising board at a bus stop is misinterpreted as a street sign.\\
\noindent\textbf{Improving performance in basic vision tasks.}  
As noted by \citet{xie2025vlms}, detection performance drops significantly when exact object locations are not provided.  
In the Challenge benchmark QA set, the location for detection was given.  
However, during real driving test time, a VLM agent must be able to perform detection autonomously.  
We also found that the VLM could not identify vanishing points on its own.  
This suggests that the VLM agent does not fully perceive the 3D environment.  
Taken together, current VLM-based driving agents struggle to achieve scene understanding, explanation, and planning simultaneously.  
Considering the goal of using VLM agents instead of end-to-end planning models, improving multi-task ability is essential.  

\noindent\textbf{Reinforcing VLMs with LiDAR Data.}  
In driving scene data collection, sensors such as LiDAR are indispensable.  
They are not only used during inference but also necessary for 3D annotation.  
Nevertheless, models like VLMs are not naturally compatible with LiDAR inputs.  
Therefore, improving the image features of VLMs through LiDAR to make them more accurate and effective will be a key research direction in the future.  

\noindent\textbf{The need for improved MLLM-based VQA metrics.}  
There is a need to refine QA-based scene understanding metrics for driving-specific settings, extending the ideas proposed in \citet{chen2024mllm}.  
In practice, current metric designs sometimes behave as if they evaluate correctness without actually inspecting the image. 
As discussed in~\cref{sec:difficulty_scoring}, there is also an issue where longer output texts tend to receive higher evaluation scores.  
Therefore, future research should focus on developing fair evaluation metrics for assessing the scene understanding ability of VLMs using MLLMs.

\section{Conclusion}
\label{conclusion}
We have presented a novel system for high-level vision-language question answering in autonomous driving. By combining multi-view and temporal perception inputs with sophisticated prompt engineering (chain-of-thought, few-shot demonstrations, metadata grounding, and self-consistency), our method achieves substantially higher accuracy than baselines. Our experiments show end-to-end gains of over 14 points in overall score (from 52.82\% to 67.37\%). Importantly, the system maintains robustness under severe visual corruption (≈96\% accuracy). Ablations confirm that each component contributes meaningfully to performance. 

In future work, we will explore integrating richer scene understanding (e.g. map information, traffic rules) and adaptive prompting strategies. We also aim to extend the evaluation to more diverse scenarios and to develop principled methods for VLM uncertainty estimation in safety-critical driving contexts. Our results suggest that with careful prompt and context design, pretrained VLMs can be a powerful tool for interpretable decision-making in autonomous vehicles.

\newpage
\newpage
{
\small
\bibliographystyle{ieeenat_fullname}
\bibliography{main}

\begin{thebibliography}{26}
\providecommand{\natexlab}[1]{#1}
\providecommand{\url}[1]{\texttt{#1}}
\expandafter\ifx\csname urlstyle\endcsname\relax
  \providecommand{\doi}[1]{doi: #1}\else
  \providecommand{\doi}{doi: \begingroup \urlstyle{rm}\Url}\fi

\bibitem[Caesar et~al.(2020)Caesar, Bankiti, Lang, Vora, Liong, Xu, Krishnan, Pan, Baldan, and Beijbom]{caesar2020nuscenes}
Holger Caesar, Varun Bankiti, Alex~H Lang, Sourabh Vora, Venice~Erin Liong, Qiang Xu, Anush Krishnan, Yu Pan, Giancarlo Baldan, and Oscar Beijbom.
\newblock nuscenes: A multimodal dataset for autonomous driving.
\newblock In \emph{Proceedings of the IEEE/CVF conference on computer vision and pattern recognition}, pages 11621--11631, 2020.

\bibitem[Chen et~al.(2024{\natexlab{a}})Chen, Chen, Zhang, Wang, Liu, Zhou, Zhang, Wan, Zhou, and Sun]{chen2024mllm}
Dongping Chen, Ruoxi Chen, Shilin Zhang, Yaochen Wang, Yinuo Liu, Huichi Zhou, Qihui Zhang, Yao Wan, Pan Zhou, and Lichao Sun.
\newblock Mllm-as-a-judge: Assessing multimodal llm-as-a-judge with vision-language benchmark.
\newblock In \emph{Forty-first International Conference on Machine Learning}, 2024{\natexlab{a}}.

\bibitem[Chen et~al.(2024{\natexlab{b}})Chen, Ma, Zhang, Xu, Qian, Yang, Fouhey, and Chai]{chen2024multi}
Xuweiyi Chen, Ziqiao Ma, Xuejun Zhang, Sihan Xu, Shengyi Qian, Jianing Yang, David Fouhey, and Joyce Chai.
\newblock Multi-object hallucination in vision language models.
\newblock \emph{Advances in Neural Information Processing Systems}, 37:\penalty0 44393--44418, 2024{\natexlab{b}}.

\bibitem[Deruyttere et~al.(2019)Deruyttere, Vandenhende, Grujicic, Van~Gool, and Moens]{deruyttere2019talk2car}
Thierry Deruyttere, Simon Vandenhende, Dusan Grujicic, Luc Van~Gool, and Marie-Francine Moens.
\newblock Talk2car: Taking control of your self-driving car.
\newblock \emph{arXiv preprint arXiv:1909.10838}, 2019.

\bibitem[Etchegaray et~al.(2025)Etchegaray, Fu, Huang, and Luo]{etchegaray2025box}
Djamahl Etchegaray, Yuxia Fu, Zi Huang, and Yadan Luo.
\newblock Box-qaymo: Box-referring vqa dataset for autonomous driving.
\newblock \emph{arXiv preprint arXiv:2507.00525}, 2025.

\bibitem[Guan et~al.(2024)Guan, Liu, Wu, Xian, Li, Liu, Wang, Chen, Huang, Yacoob, et~al.]{guan2024hallusionbench}
Tianrui Guan, Fuxiao Liu, Xiyang Wu, Ruiqi Xian, Zongxia Li, Xiaoyu Liu, Xijun Wang, Lichang Chen, Furong Huang, Yaser Yacoob, et~al.
\newblock Hallusionbench: an advanced diagnostic suite for entangled language hallucination and visual illusion in large vision-language models.
\newblock In \emph{Proceedings of the IEEE/CVF Conference on Computer Vision and Pattern Recognition}, pages 14375--14385, 2024.

\bibitem[Guo et~al.(2024)Guo, Fan, Lu, Sakaridis, and Van~Gool]{guo2024vanishing}
Diandian Guo, Deng-Ping Fan, Tongyu Lu, Christos Sakaridis, and Luc Van~Gool.
\newblock Vanishing-point-guided video semantic segmentation of driving scenes.
\newblock In \emph{Proceedings of the IEEE/CVF Conference on Computer Vision and Pattern Recognition}, pages 3544--3553, 2024.

\bibitem[Hu et~al.(2024)Hu, Shi, Fu, Roth, Ostendorf, Zettlemoyer, Smith, and Krishna]{hu2024visual}
Yushi Hu, Weijia Shi, Xingyu Fu, Dan Roth, Mari Ostendorf, Luke Zettlemoyer, Noah~A Smith, and Ranjay Krishna.
\newblock Visual sketchpad: Sketching as a visual chain of thought for multimodal language models.
\newblock \emph{Advances in Neural Information Processing Systems}, 37:\penalty0 139348--139379, 2024.

\bibitem[Hwang et~al.(2024)Hwang, Xu, Lin, Hung, Ji, Choi, Huang, He, Covington, Sapp, et~al.]{hwang2024emma}
Jyh-Jing Hwang, Runsheng Xu, Hubert Lin, Wei-Chih Hung, Jingwei Ji, Kristy Choi, Di Huang, Tong He, Paul Covington, Benjamin Sapp, et~al.
\newblock Emma: End-to-end multimodal model for autonomous driving.
\newblock \emph{arXiv preprint arXiv:2410.23262}, 2024.

\bibitem[Kim et~al.(2018)Kim, Rohrbach, Darrell, Canny, and Akata]{kim2018textual}
Jinkyu Kim, Anna Rohrbach, Trevor Darrell, John Canny, and Zeynep Akata.
\newblock Textual explanations for self-driving vehicles.
\newblock In \emph{Proceedings of the European conference on computer vision (ECCV)}, pages 563--578, 2018.

\bibitem[Lee et~al.(2025)Lee, Choi, Kang, Kim, Park, and Shim]{lee20253d}
Seonho Lee, Jiho Choi, Inha Kang, Jiwook Kim, Junsung Park, and Hyunjung Shim.
\newblock 3d-aware vision-language models fine-tuning with geometric distillation.
\newblock \emph{arXiv preprint arXiv:2506.09883}, 2025.

\bibitem[Li et~al.(2025)Li, Zhang, Zhang, Wu, Tian, Sun, Lu, Min, Liu, Lin, et~al.]{li2025r}
Chunyi Li, Jianbo Zhang, Zicheng Zhang, Haoning Wu, Yuan Tian, Wei Sun, Guo Lu, Xiongkuo Min, Xiaohong Liu, Weisi Lin, et~al.
\newblock R-bench: Are your large multimodal model robust to real-world corruptions?
\newblock \emph{IEEE Journal of Selected Topics in Signal Processing}, 2025.

\bibitem[Liang et~al.(2025)Liang, Li, Fan, Li, Nguyen, Cobbina, Bhardwaj, Chen, Liu, and Zhou]{liang2025colorbench}
Yijun Liang, Ming Li, Chenrui Fan, Ziyue Li, Dang Nguyen, Kwesi Cobbina, Shweta Bhardwaj, Jiuhai Chen, Fuxiao Liu, and Tianyi Zhou.
\newblock Colorbench: Can vlms see and understand the colorful world? a comprehensive benchmark for color perception, reasoning, and robustness.
\newblock \emph{arXiv preprint arXiv:2504.10514}, 2025.

\bibitem[Liu et~al.(2023)Liu, Lin, Hewitt, Paranjape, Bevilacqua, Petroni, and Liang]{liu2023lost}
Nelson~F Liu, Kevin Lin, John Hewitt, Ashwin Paranjape, Michele Bevilacqua, Fabio Petroni, and Percy Liang.
\newblock Lost in the middle: How language models use long contexts.
\newblock \emph{arXiv preprint arXiv:2307.03172}, 2023.

\bibitem[Marcu et~al.(2024)Marcu, Chen, H{\"u}nermann, Karnsund, Hanotte, Chidananda, Nair, Badrinarayanan, Kendall, Shotton, et~al.]{marcu2024lingoqa}
Ana-Maria Marcu, Long Chen, Jan H{\"u}nermann, Alice Karnsund, Benoit Hanotte, Prajwal Chidananda, Saurabh Nair, Vijay Badrinarayanan, Alex Kendall, Jamie Shotton, et~al.
\newblock Lingoqa: Visual question answering for autonomous driving.
\newblock In \emph{European Conference on Computer Vision}, pages 252--269. Springer, 2024.

\bibitem[Qian et~al.(2024)Qian, Chen, Zhuo, Jiao, and Jiang]{qian2024nuscenes}
Tianwen Qian, Jingjing Chen, Linhai Zhuo, Yang Jiao, and Yu-Gang Jiang.
\newblock Nuscenes-qa: A multi-modal visual question answering benchmark for autonomous driving scenario.
\newblock In \emph{Proceedings of the AAAI Conference on Artificial Intelligence}, pages 4542--4550, 2024.

\bibitem[Sachdeva et~al.(2024)Sachdeva, Agarwal, Chundi, Roelofs, Li, Kochenderfer, Choi, and Dariush]{sachdeva2024rank2tell}
Enna Sachdeva, Nakul Agarwal, Suhas Chundi, Sean Roelofs, Jiachen Li, Mykel Kochenderfer, Chiho Choi, and Behzad Dariush.
\newblock Rank2tell: A multimodal driving dataset for joint importance ranking and reasoning.
\newblock In \emph{Proceedings of the IEEE/CVF winter conference on applications of computer vision}, pages 7513--7522, 2024.

\bibitem[Shao et~al.(2024)Shao, Hu, Wang, Song, Waslander, Liu, and Li]{shao2024lmdrive}
Hao Shao, Yuxuan Hu, Letian Wang, Guanglu Song, Steven~L Waslander, Yu Liu, and Hongsheng Li.
\newblock Lmdrive: Closed-loop end-to-end driving with large language models.
\newblock In \emph{Proceedings of the IEEE/CVF Conference on Computer Vision and Pattern Recognition}, pages 15120--15130, 2024.

\bibitem[Sima et~al.(2024)Sima, Renz, Chitta, Chen, Zhang, Xie, Bei{\ss}wenger, Luo, Geiger, and Li]{sima2024drivelm}
Chonghao Sima, Katrin Renz, Kashyap Chitta, Li Chen, Hanxue Zhang, Chengen Xie, Jens Bei{\ss}wenger, Ping Luo, Andreas Geiger, and Hongyang Li.
\newblock Drivelm: Driving with graph visual question answering.
\newblock In \emph{European conference on computer vision}, pages 256--274. Springer, 2024.

\bibitem[Teney et~al.(2017)Teney, Liu, and van Den~Hengel]{teney2017graph}
Damien Teney, Lingqiao Liu, and Anton van Den~Hengel.
\newblock Graph-structured representations for visual question answering.
\newblock In \emph{Proceedings of the IEEE conference on computer vision and pattern recognition}, pages 1--9, 2017.

\bibitem[Tian et~al.(2024)Tian, Gu, Li, Liu, Wang, Zhao, Zhan, Jia, Lang, and Zhao]{tian2024drivevlm}
Xiaoyu Tian, Junru Gu, Bailin Li, Yicheng Liu, Yang Wang, Zhiyong Zhao, Kun Zhan, Peng Jia, Xianpeng Lang, and Hang Zhao.
\newblock Drivevlm: The convergence of autonomous driving and large vision-language models.
\newblock \emph{arXiv preprint arXiv:2402.12289}, 2024.

\bibitem[Wang et~al.(2024)Wang, Leroy, Cabon, Chidlovskii, and Revaud]{wang2024dust3r}
Shuzhe Wang, Vincent Leroy, Yohann Cabon, Boris Chidlovskii, and Jerome Revaud.
\newblock Dust3r: Geometric 3d vision made easy.
\newblock In \emph{Proceedings of the IEEE/CVF Conference on Computer Vision and Pattern Recognition}, pages 20697--20709, 2024.

\bibitem[Wang et~al.(2025)Wang, Yu, Jiang, Lan, Shi, Chang, Kautz, Li, and Alvarez]{wang2025omnidrive}
Shihao Wang, Zhiding Yu, Xiaohui Jiang, Shiyi Lan, Min Shi, Nadine Chang, Jan Kautz, Ying Li, and Jose~M Alvarez.
\newblock Omnidrive: A holistic vision-language dataset for autonomous driving with counterfactual reasoning.
\newblock In \emph{Proceedings of the Computer Vision and Pattern Recognition Conference}, pages 22442--22452, 2025.

\bibitem[Wang et~al.(2022)Wang, Wei, Schuurmans, Le, Chi, Narang, Chowdhery, and Zhou]{wang2022self}
Xuezhi Wang, Jason Wei, Dale Schuurmans, Quoc Le, Ed Chi, Sharan Narang, Aakanksha Chowdhery, and Denny Zhou.
\newblock Self-consistency improves chain of thought reasoning in language models.
\newblock \emph{arXiv preprint arXiv:2203.11171}, 2022.

\bibitem[Xie et~al.(2025)Xie, Kong, Dong, Sima, Zhang, Chen, Liu, and Pan]{xie2025vlms}
Shaoyuan Xie, Lingdong Kong, Yuhao Dong, Chonghao Sima, Wenwei Zhang, Qi~Alfred Chen, Ziwei Liu, and Liang Pan.
\newblock Are vlms ready for autonomous driving? an empirical study from the reliability, data, and metric perspectives.
\newblock \emph{arXiv preprint arXiv:2501.04003}, 2025.

\bibitem[Xu et~al.(2024)Xu, Zhang, Xie, Zhao, Guo, Wong, Li, and Zhao]{xu2024drivegpt4}
Zhenhua Xu, Yujia Zhang, Enze Xie, Zhen Zhao, Yong Guo, Kwan-Yee~K Wong, Zhenguo Li, and Hengshuang Zhao.
\newblock Drivegpt4: Interpretable end-to-end autonomous driving via large language model.
\newblock \emph{IEEE Robotics and Automation Letters}, 2024.

\end{thebibliography}
}

\end{document}